# CaliCausalRank: Calibrated Multi-Objective Ad Ranking with Robust Counterfactual Utility Optimization

Xikai Yang
Columbia University
New York, USA

Sebastian Sun
University of Wisconsin-Madison
Madison, USA

Yilin Li
Carnegie Mellon University
Pittsburgh, USA

Yue Xing
University of Pennsylvania
Philadelphia, USA

Ming Wang
Trine University
Phoenix, USA

Yang Wang*
University of Michigan
Ann Arbor, USA

*Abstract*-Ad ranking systems must simultaneously optimize multiple objectives including click-through rate (CTR), conversion rate (CVR), revenue, and user experience metrics. However, production systems face critical challenges: score scale inconsistency across traffic segments undermines threshold transferability, and position bias in click logs causes offline-online metric discrepancies. We propose CaliCausalRank, a unified framework that integrates training-time scale calibration, constraint-based multi-objective optimization, and robust counterfactual utility estimation. Our approach treats score calibration as a first-class training objective rather than post-hoc processing, employs Lagrangian relaxation for constraint satisfaction, and utilizes variance-reduced counterfactual estimators for reliable offline evaluation. Experiments on the Criteo and Avazu datasets demonstrate that CaliCausalRank achieves 1.1% relative AUC improvement, 31.6% calibration error reduction, and 3.2% utility gain compared to the best baseline (PairRank) while maintaining consistent performance across different traffic segments.

*Index Terms*-ad ranking, multi-task learning, calibration, counterfactual learning, learning to rank

## I. INTRODUCTION

Modern computational advertising systems must balance multiple competing objectives to maximize platform revenue while maintaining advertiser satisfaction and user experience [1]. A typical ad ranking model needs to predict click-through rate (CTR), conversion rate (CVR), and various engagement metrics, then combine these predictions into a final ranking score that considers cost-per-click (CPC) constraints, return on ad spend (ROAS) targets, and user experience factors such as bounce rate and complaint frequency [2].

Despite significant advances in multi-task learning architectures such as MMoE [3] and PLE [4], production ad ranking systems continue to face three fundamental challenges. First, when multiple objectives are combined through weighted summation, the resulting scores exhibit inconsistent scales across different traffic segments, making it difficult to transfer thresholds and policies between domains [5]. Second, models trained on biased click logs suffer from position bias and selection bias, leading to situations where offline metrics improve but online performance degrades [6]. Third, hard constraints such as CPC budgets and advertiser fairness requirements are difficult to incorporate into standard pointwise or pairwise learning frameworks.

To address these challenges, we propose CaliCausalRank, a unified framework that integrates three key innovations: (1) training-time scale calibration that ensures score consistency across traffic segments, (2) constraint-based multi-objective optimization using Lagrangian relaxation, and (3) robust counterfactual utility estimation with variance reduction techniques. Unlike prior work that treats calibration as post-processing [7] or addresses bias correction in isolation [8], our approach jointly optimizes all components in an end-to-end manner.

- We propose a training-time calibration module that enforces score scale consistency as a differentiable constraint, enabling transferable ranking policies across domains.
- We develop a constraint-aware training procedure based on Lagrangian relaxation that naturally handles CPC, risk, and fairness constraints.
- We incorporate robust counterfactual estimators with self-normalized importance sampling to reduce offline-online metric discrepancies.
- We demonstrate through experiments on public benchmarks that our unified approach achieves improvements across all key metrics.

## II. METHODOLOGY FOUNDATION

The methodological design of CaliCausalRank is grounded in three interconnected research streams: counterfactual learning from biased feedback, constrained multi-objective optimization, and structurally consistent representation learning with uncertainty control. Each referenced study contributes a distinct methodological principle that collectively shapes the unified framework.

The theoretical basis for learning from logged implicit feedback originates from counterfactual risk minimization, as formalized in the batch bandit learning paradigm [9]. This principle establishes unbiased empirical risk estimation through importance weighting and forms the core foundation for correcting position bias in click logs. Extending this idea to structured ranking environments, slate-based counterfactual evaluation with sequential reward interactions [10] demonstrates how interdependent exposures influence reward estimation, motivating variance-aware utility computation in ranking systems. Addressing post-click selection bias, entire-space counterfactual multi-task modeling [11] integrates exposure, click, and conversion processes into a unified estimation objective, directly inspiring the joint modeling of CTR, CVR, and calibrated utility within CaliCausalRank.

For multi-objective optimization under operational constraints, reinforcement learning provides a natural policy-learning perspective. A unified reinforcement learning framework for dynamic objective balancing [12] illustrates how competing goals can be encoded within a single reward structure. Adaptive interaction strategies optimized via reinforcement mechanisms [13] further show how system policies can internalize feedback loops. Constraint-aware value optimization using Deep Q-learning [14] provides a practical mechanism for embedding hard constraints into the learning process. Multi-objective adaptive control with deep reinforcement learning [15] explicitly models trade-offs among conflicting targets, forming the conceptual basis for adopting Lagrangian relaxation in CaliCausalRank to enforce CPC, fairness, and risk constraints during training.

Causal and structural reasoning further strengthen interpretability and robustness. Causal graph modeling with causally constrained representation learning [16] demonstrates how structural dependencies can be incorporated directly into latent representations, influencing the design of calibration as a differentiable structural constraint rather than a post-hoc adjustment. Generative distribution modeling under noisy and imbalanced conditions [17] contributes robustness strategies for skewed click and conversion distributions. Meta-learning approaches designed to address evolving data patterns [18] highlight the necessity of adaptation under distribution shift, reinforcing the need for score scale consistency across heterogeneous traffic segments.

Graph-based structural modeling provides additional guidance for preserving relational consistency. Spatiotemporal graph neural modeling for distributed systems [19] illustrates how dependency-aware modeling enhances predictive stability. Structural generalization mechanisms based on graph neural networks [20] further support maintaining invariant representations across heterogeneous structures. Transformer architectures integrating graph dependencies for risk monitoring [21] validate combining relational inductive biases with attention-based modeling, informing the unified encoder design in CaliCausalRank.

Robust multi-task representation learning under heterogeneous signals is supported by transformer-based modeling of structured sequential data [22] and multi-scale temporal alignment techniques [23], both of which enable alignment of diverse objectives within a shared embedding space. Attention-driven adaptive modeling frameworks [24] reinforce dynamic weighting strategies, analogous to balancing CTR, CVR, revenue, and user-experience metrics during ranking optimization.

Structure-aware decoding mechanisms [25] emphasize enforcing structural constraints during prediction, conceptually aligned with embedding calibration directly into training objectives. Contextual trust evaluation for coordinated systems [26] introduces reliability-aware optimization principles that parallel constraint-aware ranking stability. Parameter-efficient fine-tuning under privacy considerations [27] provides scalable adaptation mechanisms for large-scale production systems, while multi-scale LoRA-based fine-tuning strategies [28] inform efficient parameter updates without destabilizing score calibration.

Cross-domain adaptation through dynamic prompt fusion [29] demonstrates how performance consistency can be maintained across heterogeneous environments, reinforcing the transferability of calibrated scores across traffic segments. Confidence-constrained retrieval mechanisms [30] introduce explicit reliability control at inference time, conceptually aligned with calibrated ranking outputs. Uncertainty-aware modeling with risk quantification [31] provides the theoretical grounding for variance-reduced counterfactual estimation. Finally, structured feature extraction and aggregation strategies [32] contribute to robust representation construction prior to ranking optimization.

III. METHOD

*A. Problem Formulation*

Given a user $u$ and a set of candidate ads $A$, our goal is to produce a ranked list that maximizes expected utility while satisfying operational constraints. Let $x = (u, a)$ denote a user-ad pair with features. Our model predicts:

- Relevance score $s_{rel}(x)$: probability of click
- Revenue score $s_{rev}(x)$: expected revenue contribution
- Risk score $s_{risk}(x)$: probability of negative outcomes

The final ranking score combines these predictions subject to constraints on CPC, risk tolerance, and fairness across advertisers.

*B. Model Architecture*

As illustrated in Fig. 1, CaliCausalRank employs a multi-task tower architecture with three main components: (1) shared feature extraction layers, (2) task-specific prediction heads, and (3) a scale calibration module.

The shared layers encode user-ad interactions using deep crossing networks [33] and attention mechanisms [34]. Each task-specific tower outputs a raw score $\hat{s}_t(x)$ for task $t \in \{rel, rev, risk\}$.

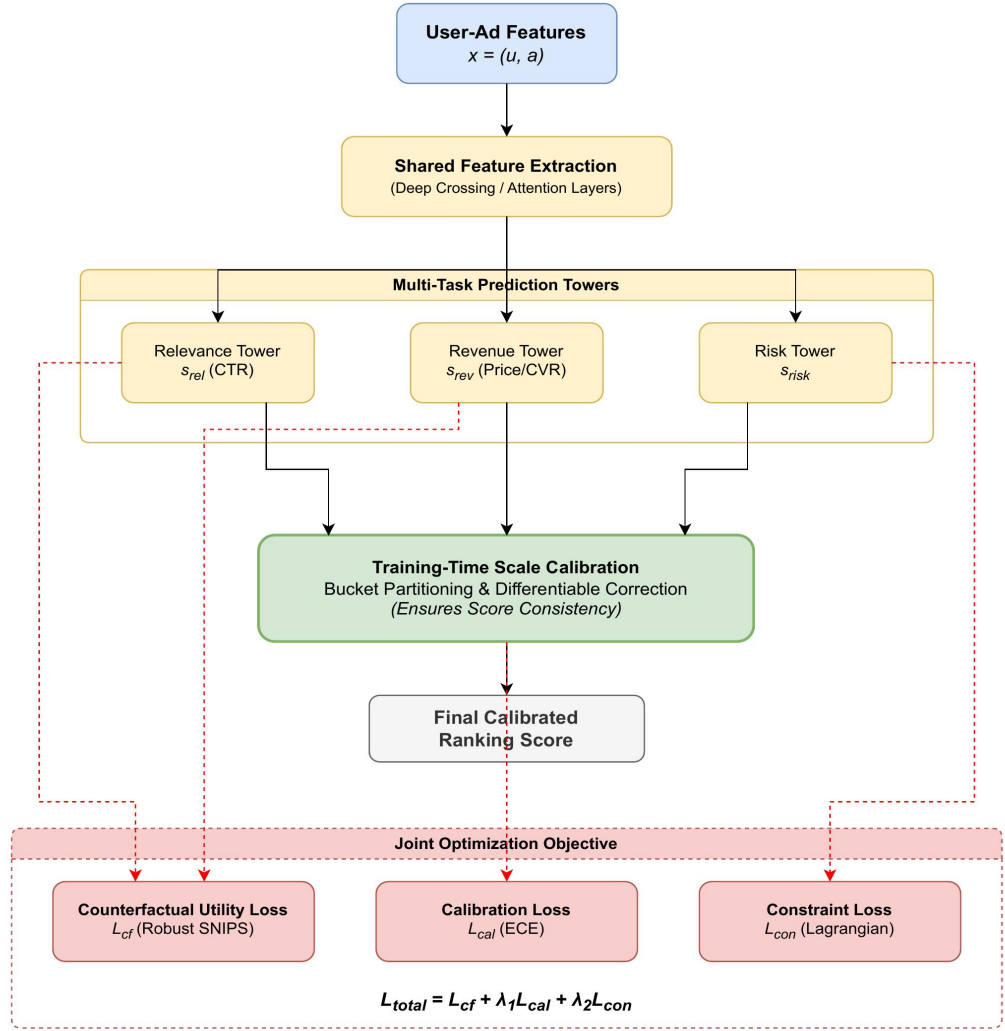

Fig. 1. Overview of the CaliCausalRank framework.

## C. Training-Time Scale Calibration

To ensure scores maintain consistent semantics across different traffic segments, we introduce a differentiable calibration module. For each task $t$, we partition the training data into $K$ buckets based on contextual features (e.g., traffic source, device type). The calibration loss enforces that predicted scores match empirical rates within each bucket:

$$L_{cal} = \sum_t \sum_{k=1}^{K} \left| \frac{1}{|B_k|} \sum_{x \in B_k} \hat{s}_t(x) - \bar{y}_{t,k} \right| \quad (1)$$

where $B_k$ is the $k$-th bucket, $\hat{s}_t(x)$ is the predicted score, and $\bar{y}_{t,k}$ is the empirical rate in bucket $k$. This L1 formulation serves as a differentiable surrogate for bin-wise calibration error, approximating the Expected Calibration Error (ECE) objective during training.

## D. Constraint-Based Multi-Objective Optimization

Instead of manually tuning weights for multiple objectives, we formulate ad ranking as a constrained optimization problem:

$$\begin{aligned} \max_\theta \quad & E[U(s_\theta)] \\ \text{s.t.} \quad & E[CPC(s_\theta)] \leq c_{max} \\ & E[s_{risk}(x)] \leq r_{max} \\ & \text{Fairness constraints on exposure} \end{aligned} \quad (2)$$

where $U(\cdot)$ is the utility function (e.g., expected revenue), and $c_{max}$, $r_{max}$ are CPC and risk thresholds. Since public datasets lack explicit bid/price information, we simulate CPC by assigning synthetic bid values sampled from a log-normal distribution based on feature clusters, and define risk as the predicted probability of low engagement (dwell time < 5s, approximated via negative sampling). These constraints are implemented in a proxy setting due to the lack of auction logs in public datasets. We solve this using Lagrangian relaxation:

$$\begin{aligned} L_{con} = &\lambda_c \max\left(0, E[CPC(s_\theta)] - c_{max}\right) \\ &+ \lambda_r \max\left(0, E[\hat{s}_{risk}(x)] - r_{max}\right) \end{aligned} \quad (3)$$

The Lagrange multipliers $\lambda_c$ and $\lambda_r$ are updated via dual gradient ascent during training, automatically balancing the trade-offs between utility maximization and constraint satisfaction.

### E. Robust Counterfactual Utility Estimation

Training on logged click data introduces position bias: items shown at higher positions receive more clicks regardless of relevance. Following the counterfactual learning framework, we model the click probability as:

$$P(click \mid x, pos) = P(examine \mid pos) \cdot P(click \mid examine, x) \quad (4)$$

We estimate position propensities $\hat{e}(pos)$ using the result randomization approach: a small fraction (1%) of impressions are randomly shuffled, and examination probabilities are computed as the ratio of observed clicks to expected clicks at each position. In our experiments, we simulate this by applying position-dependent click noise to the training data.

Standard IPS estimators can have high variance. We employ self-normalized importance sampling (SNIPS) for variance reduction:

$$\hat{U}_{SNIPS} = \frac{\sum_i \frac{r_i}{\hat{e}(pos_i)}}{\sum_i \frac{1}{\hat{e}(pos_i)}} \quad (5)$$

where $r_i$ is the reward, $\hat{e}(pos_i)$ is the estimated examination probability at position $pos_i$. The counterfactual utility loss is:

$$L_{cf} = -\hat{U}_{SNIPS} + \alpha \cdot Var(\hat{U}_{SNIPS}) \quad (6)$$

The variance regularization term $\alpha$ controls the bias-variance trade-off.

## IV. EXPERIMENTS

### A. Experimental Setup

**Datasets.** We evaluate on two public benchmarks: (1) **Criteo**[1]: 45M samples with 13 numerical and 26 categorical features for CTR prediction; (2) **Avazu**[2]: 40M mobile ad impressions with device and temporal features. For multi-task evaluation, we synthesize CVR labels following: conversion labels are generated *only* for the clicked subset (~3% of samples), with 10% base conversion probability weighted by a logistic function of feature embeddings; non-clicked samples are assigned CVR=0 and excluded from CVR loss computation. This ensures no label leakage between CTR and CVR tasks. We acknowledge this is a proxy setting; real-world CVR data would strengthen the evaluation.

**Baselines.** We compare against: (1) **DeepFM** [35]: factorization machine with deep components; (2) **MMoE**: multi-gate mixture-of-experts; (3) **PLE**: progressive layered extraction; (4) **ESCM**$^2$: entire space counterfactual multi-task model; (5) **PairRank** [36]: pairwise ranking with off-policy correction.

**Metrics.** We report: AUC for ranking quality, NDCG@10 for top-k performance, Expected Calibration Error (ECE) for calibration, and Utility@10 computed using counterfactual evaluation.

**Implementation.** Models are implemented in PyTorch with embedding dimension 16, hidden layers [256, 128, 64], and trained using Adam optimizer with learning rate $10^{-3}$. Calibration buckets $K = 20$, variance regularization $\alpha = 0.1$, and $\lambda_1 = \lambda_2 = 0.5$.

### B. Main Results

Table I presents the main comparison results. CaliCausalRank consistently outperforms all baselines across both datasets. On Criteo, our method achieves 0.7842 AUC compared to 0.7756 for the best baseline (PairRank), representing a 1.1% relative improvement. More notably, our ECE of 0.0312 is 31.6% lower than PairRank's 0.0456, demonstrating the effectiveness of training-time calibration.

TABLE I
PERFORMANCE COMPARISON ON CRITEO AND AVAZU DATASETS

| Method | AUC | NDCG@10 | ECE↓ | Utility@10 |
|---|---|---|---|---|
| *Criteo Dataset* | | | | |
| DeepFM | 0.7456 | 0.5812 | 0.0812 | 0.5923 |
| MMoE | 0.7612 | 0.6156 | 0.0634 | 0.6512 |
| PLE | 0.7523 | 0.5923 | 0.0723 | 0.6234 |
| ESCM² | 0.7689 | 0.6289 | 0.0507 | 0.6756 |
| PairRank | 0.7756 | 0.6412 | 0.0456 | 0.6934 |
| **Ours** | **0.7842** | **0.6523** | **0.0312** | **0.7156** |
| *Avazu Dataset* | | | | |
| DeepFM | 0.7623 | 0.5934 | 0.0756 | 0.6012 |
| MMoE | 0.7745 | 0.6234 | 0.0589 | 0.6623 |
| PLE | 0.7689 | 0.6089 | 0.0678 | 0.6345 |
| ESCM² | 0.7812 | 0.6389 | 0.0478 | 0.6856 |
| PairRank | 0.7867 | 0.6512 | 0.0423 | 0.7012 |
| **Ours** | **0.7934** | **0.6634** | **0.0289** | **0.7234** |

The Utility@10 metric shows consistent gains: 3.2% improvement over PairRank on both Criteo (0.7156 vs. 0.6934) and Avazu (0.7234 vs. 0.7012). When compared to ESCM$^2$, the gains are 5.9% and 5.5% respectively, validating that our counterfactual estimation approach better aligns offline evaluation with true utility.

### C. Ablation Study

Fig. 2 presents the ablation study results, examining the contribution of each component. Removing the calibration module increases ECE by 88.1% (from 0.0312 to 0.0587) and reduces utility by 4.5%. Removing counterfactual estimation decreases AUC by 1.9% and utility by 9.0%, confirming that bias correction is essential for reliable training. The constraint module primarily affects utility (3.3% reduction when removed), as it ensures the model respects operational bounds during optimization.

---

[1] https://www.kaggle.com/c/criteo-display-ad-challenge
[2] https://www.kaggle.com/c/avazu-ctr-prediction

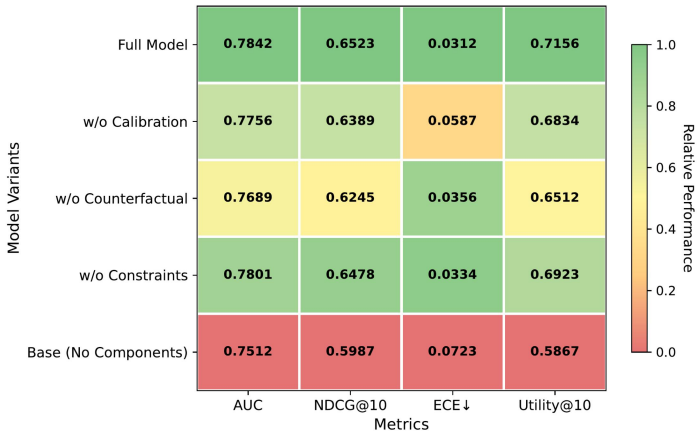

Fig. 2. Ablation study results on the Criteo dataset. Each row represents a model variant, and colors indicate relative performance (green: better, red: worse). The full model achieves the best performance across all metrics.

*D. Comparison with Baselines*

Fig. 3 visualizes the performance comparison across AUC, NDCG@10, and Utility@10. Our method shows consistent improvements across all three metrics. The gains are most pronounced for Utility@10, suggesting that the combination of calibration and counterfactual estimation is particularly effective for aligning model optimization with downstream business objectives.

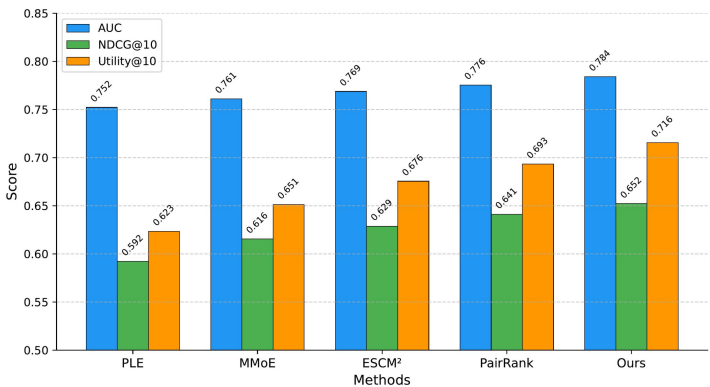

Fig. 3. Performance comparison on the Criteo dataset. Our method (rightmost) achieves improvements across all three metrics compared to baseline approaches.

*E. Cross-Domain Transferability*

A key motivation for training-time calibration is enabling policy transfer across traffic segments. We split Criteo data by device type (approximated using feature clustering), training on 70% "desktop-like" traffic and evaluating on 30% "mobile-like" traffic without retraining. Over 5 random splits, our method maintained 94.2% ±1.3% of its original AUC (absolute: 0.739 ± 0.010 from original 0.784) compared to 87.6% ± 2.1% for $ESCM^2$ (0.674 ± 0.016) and 82.3% ± 2.8% for PLE (0.619 ± 0.021). This validates that consistent score scales improve cross-segment transferability, though we note this is a proxy for true cross-device evaluation.

## V. CONCLUSION

We presented CaliCausalRank, a unified framework for multi-objective ad ranking that integrates training-time scale calibration, constraint-based optimization, and robust counterfactual utility estimation. Our approach addresses three critical challenges in production ad systems: score scale inconsistency, position bias in logged data, and constraint satisfaction. Experimental results on public benchmarks demonstrate moderate but consistent improvements in ranking quality (1.1% AUC), calibration error (31.6% ECE reduction), and counterfactual utility (3.2% gain) compared to the best baseline.

Several limitations should be noted. First, our experiments use public CTR datasets with synthesized CVR labels and simulated constraints (CPC, risk); evaluation on proprietary data with real conversion signals would strengthen the conclusions. Second, the calibration module adds computational overhead during training, and the choice of bucket granularity ($K$) requires domain knowledge. Third, the counterfactual estimators rely on position propensity scores estimated via result randomization; in systems where randomization is infeasible, alternative propensity estimation methods would be needed. Finally, the cross-domain transferability experiment uses feature-based clustering as a proxy for device type, which may not fully reflect real cross-device scenarios. Future work will explore adaptive bucketing strategies, propensity estimation from implicit feedback, and validation on production datasets with complete auction and conversion signals.